\documentclass[letterpaper]{article} 
\usepackage{aaai2027}

\usepackage[hyphens]{url}  
\usepackage{graphicx} 
\usepackage{natbib}  
\usepackage{caption} 
\usepackage[T1]{fontenc}
\usepackage{booktabs}       
\usepackage{amsfonts}       
\usepackage{nicefrac}       
\usepackage{microtype}      
\usepackage[table]{xcolor}  
\usepackage{subcaption}     
\usepackage{amsmath,amssymb}
\usepackage{multirow}       
\newcommand{\legendbox}[1]{{\color{#1}\rule{3ex}{2ex}}}
\definecolor{rankfirst}{HTML}{E78B5F}  
\definecolor{ranksecond}{HTML}{F3BEA2} 
\definecolor{rankthird}{HTML}{F9E3D6}  

\newcommand{\first}[1]{\colorbox{rankfirst}{\strut\hspace{0.30em}#1\hspace{0.30em}}}
\newcommand{\second}[1]{\colorbox{ranksecond}{\strut\hspace{0.30em}#1\hspace{0.30em}}}
\newcommand{\third}[1]{\colorbox{rankthird}{\strut\hspace{0.30em}#1\hspace{0.30em}}}

\title{TextAlign: Preference Alignment for Text Rendering with Hierarchical Rewards}

\author{
  Mingxuan Cui\textsuperscript{\rm 1}\equalcontrib,
  Jingpu Yang\textsuperscript{\rm 2}\equalcontrib,
  Fengxian Ji\textsuperscript{\rm 1}\equalcontrib,
  Qian Jiang\textsuperscript{\rm 3},
  Zhecheng Shi\textsuperscript{\rm 4},\\
  Jiaming Wang\textsuperscript{\rm 3},
  Zirui Song\textsuperscript{\rm 1},
  Zhuohan Xie\textsuperscript{\rm 1},
  Fajri Koto\textsuperscript{\rm 1},
  Xiuying Chen\textsuperscript{\rm 1}\thanks{Corresponding author.}
}

\affiliations{
  \textsuperscript{\rm 1}Mohamed bin Zayed University of Artificial Intelligence\\
  \textsuperscript{\rm 2}Chinese Academy of Sciences Institute of Automation\\
  \textsuperscript{\rm 3}Northeastern University\\
  \textsuperscript{\rm 4}The Hong Kong University of Science and Technology (Guangzhou)
}

\begin{document}

\maketitle

\begin{abstract}
Faithful text rendering remains a persistent weakness of large text-to-image generative models, as it requires both semantic instruction following and fine-grained glyph-level structure.
Prior methods often improve this ability through architecture-specific modules or encoder modifications, which complicate deployment across foundation models.
We study text rendering as a post-training preference-alignment problem and propose TextAlign, a non-invasive framework that keeps the generator architecture unchanged.
The key component is a hierarchical vision-language model (VLM)-based reward that decomposes rendering errors into global, word, and glyph levels, then converts binary defect judgments into a scalar preference signal.
The resulting signal supports both Group Relative Policy Optimization (GRPO) and Direct Preference Optimization (DPO).
Experiments across SD3.5-Medium, FLUX.1-dev, and Z-Image-Turbo show consistent gains in OCR-based text accuracy without degrading general generation quality. Compared with strong foundation and text-rendering baselines, including Qwen-Image, AnyText, and TextDiffuser, these results indicate that reward design offers a scalable alternative to model redesign for improving text rendering.
\end{abstract}

\section{Introduction}

\begin{figure}[!t]
  \centering
  \includegraphics[width=0.95\linewidth]{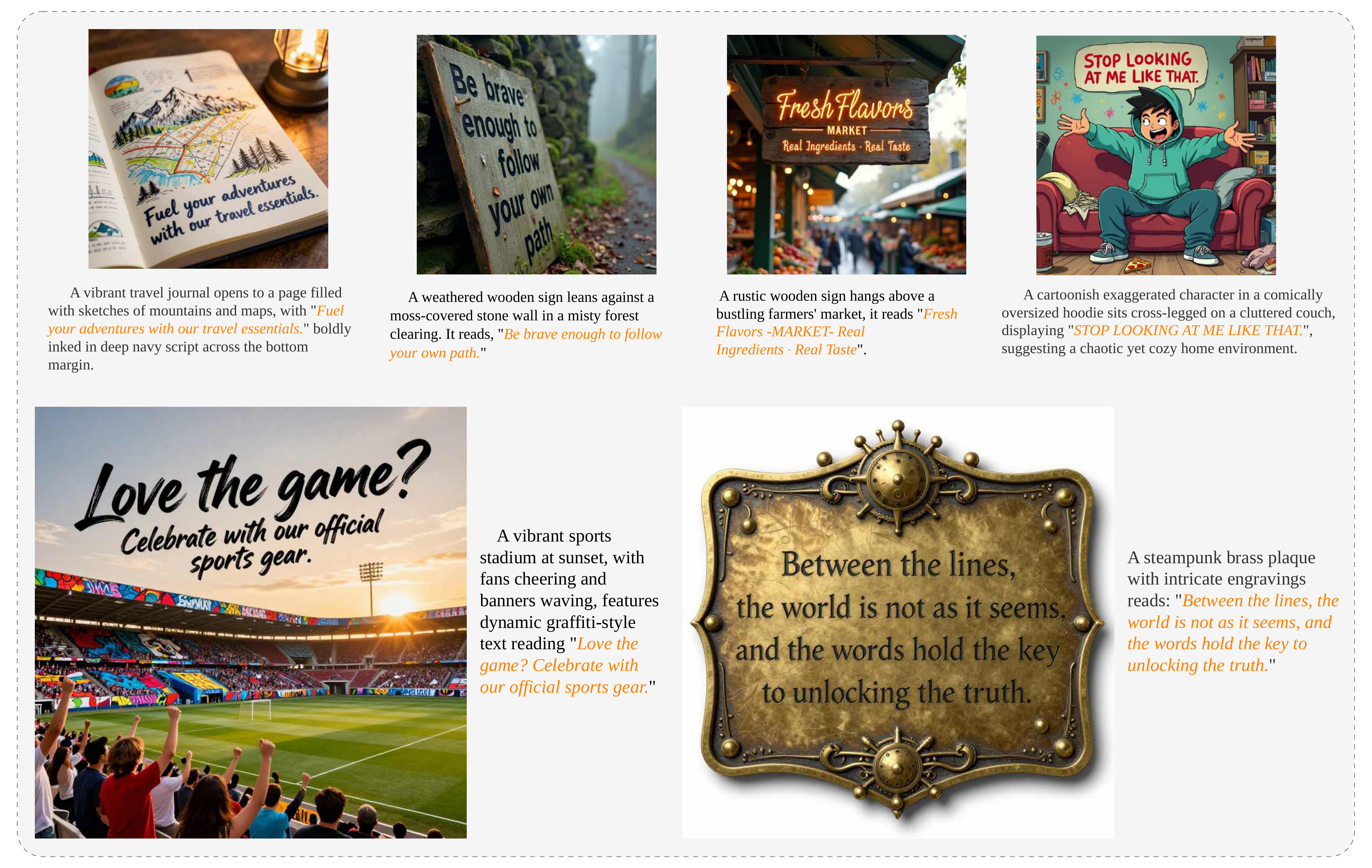}
  \caption{\textbf{Text rendering results.} Representative 720 $\times$ 720 samples generated by our aligned models. TextAlign renders legible and well-formed visual text across diverse carriers, styles, layouts, and text lengths while preserving coherent image content.}
  \label{fig:results}
\end{figure}

Large text-to-image generative models~\cite{sd3, flux, qwenimage, zimage} have made substantial progress in synthesizing high-fidelity images from open-ended natural language instructions. However, faithfully rendering text inside images remains a persistent failure mode~\cite{yang1}. Unlike generic object or style generation, visual text rendering requires the model to satisfy a discrete symbolic constraint while also producing continuous visual structure: the intended words must appear, characters must be ordered correctly, glyph shapes must be recognizable, and the text must be placed naturally within the surrounding scene. This makes text rendering a stringent test of both instruction following and fine-grained visual precision. As illustrated in Fig.~\ref{fig:results}, successful rendering must hold across varied carriers, layouts, styles, and string lengths, rather than only in simplified centered-text settings.

Existing methods largely address this problem through architecture-level interventions. Some approaches replace or augment text and image encoders to strengthen character awareness~\cite{character-aware, udifftext, glyphbyt5, vitype}, while others introduce additional control modules or glyph-level conditions to guide the spatial layout of rendered text~\cite{textdiffuser, glyphcontrol, anytext, uniglyph}. These designs have improved visual text synthesis, but they also introduce a practical limitation: the solution is often tied to a particular model architecture, conditioning interface, or control representation. As a result, transferring the same recipe to a new foundation model can require non-trivial engineering and may disturb the pretrained generative prior that gives modern models their broad visual competence.

We take a different view: text rendering can be treated as a post-training preference-alignment problem. Rather than redesigning the generator, we ask whether a frozen foundation model can be aligned toward better typographic behavior through a reward signal that accurately reflects rendering failures. This perspective is attractive because it matches the deployment pattern of modern generative models: the base architecture is kept intact, while post-training adjusts model behavior toward a desired capability. It also makes the problem model-agnostic, enabling the same alignment framework to be applied to different image generators without introducing task-specific trainable modules.

To this end, we propose \textbf{TextAlign}, a non-invasive preference-alignment framework for visual text rendering. The central technical challenge is reward design. Text rendering errors are inherently hierarchical: a generated image may contain no readable text at all, may render the wrong set of words, or may preserve the words while corrupting individual characters. A single OCR score or binary text-presence reward cannot reliably distinguish these failure modes. TextAlign therefore decomposes rendering quality into three levels: a \emph{global} level that detects missing or malformed text, a \emph{word} level that captures dropped, inserted, or substituted words, and a \emph{glyph} level that identifies character-level insertions, deletions, and substitutions. Each level is judged by a vision-language model (VLM) through structured binary defect indicators, which are aggregated into a scalar reward for preference optimization.

The resulting reward can be used by both Group Relative Policy Optimization (GRPO)~\cite{deepseekmath, dancegrpo} and Direct Preference Optimization (DPO)~\cite{dpo, diffusiondpo}. This compatibility allows TextAlign to serve as a general post-training interface rather than a method bound to a single optimizer. Importantly, because the reward is defined over generated images and target text, it does not require architectural modification, auxiliary glyph-control branches, or model-specific supervision signals.

We evaluate TextAlign on FLUX.1-dev~\cite{flux} and Z-Image-Turbo~\cite{zimage}, and compare against strong foundation-model references including SD3.5~\cite{sd3} and Qwen-Image~\cite{qwenimage}, as well as representative text-rendering systems such as TextDiffuser~\cite{textdiffuser} and AnyText~\cite{anytext}. Across OCR-based metrics, both DPO and GRPO consistently improve text rendering accuracy; on the stronger Z-Image-Turbo backbone, the GRPO-aligned model achieves the best normalized edit score, precision, recall, and F1-score among all compared methods. These gains are not obtained by sacrificing the original image prior: CLIPScore, ImageReward, PickScore, and HPSv3 remain competitive after alignment, and a human study further favors the aligned models in both text fidelity and visual integration. We also analyze robustness across text length, spatial placement, and ten visual categories, where the aligned model maintains stable performance rather than overfitting to short, centered, or category-specific text. Taken together, these results support the central claim of this paper: reliable visual text rendering can be improved through carefully designed reward modeling and post-training alignment, without changing the architecture of large image generative models.

\section{Related work}

\textbf{Diffusion models}~\cite{scoresde,ji1,yang2,ji2,yang4,yang5,lai1, kong2024hunyuanvideo, yuan2023hap} have been widely adopted in text-to-image generation.
They use a diffusion process to gradually add noise to the image for training and then reverse the process for generation.
Subsequent advances in latent-space modeling~\cite{ldm,ji3}, transformer backbones~\cite{dit}, and rectified-flow or flow-matching formulations~\cite{flowmatching,rectifiedflow} have further improved the scalability and sample quality of this paradigm.
At present, open text-to-image systems such as Stable Diffusion~3~\cite{sd3}, FLUX~\cite{flux}, Qwen-Image~\cite{qwenimage}, and Z-Image~\cite{zimage} represent the frontier of large-scale image generation.

\textbf{Text rendering} in text-to-image diffusion models has been advanced along several complementary directions.
Much prior work injects layout or glyph information through auxiliary control modules~\cite{glyphdraw, textdiffuser, glyphcontrol, anytext, textdiffuser2, chen2025infinite, brushyourtext, glyphbyt5, dreamtext, controlinfo, visualtextinwild, textmaster, fonts, uniglyph, ma2024follow}.
For example, TextDiffuser~\cite{textdiffuser} first generates character-level segmentation masks and then performs mask-conditioned generation with character-aware supervision.
Another line of work employs specialized text or image encoders~\cite{character-aware, udifftext, vitype}.
For example, UDiffText~\cite{udifftext} trains a character-aware text encoder with a codebook to replace the original text encoder.
Yet another line explores complex training strategies~\cite{designdiffusion, preciseparameterlocalization, amo}.
For example, the AMO sampler~\cite{amo,song1,meng1} alternates between ODE overshooting and noise reintroduction to introduce Langevin dynamics correction, while adaptively controlling overshooting strength via cross-attention scores.

\textbf{Preference alignment} in text-to-image diffusion models has been explored recently.
One active thread adapts reinforcement learning with group-relative credit assignment under Group Relative Policy Optimization (GRPO)~\cite{deepseekmath}; DanceGRPO~\cite{dancegrpo} brings GRPO to visual generation and reports stable scaling across diffusion and rectified-flow models under diverse reward signals.
A complementary thread avoids explicit reward modeling and instead aligns the generative policy with pairwise preferences: Diffusion-DPO~\cite{diffusiondpo} extends Direct Preference Optimization (DPO)~\cite{dpo,yang3} to diffusion by defining a likelihood-compatible objective and shows strong improvements from human comparison data.
However, these approaches typically optimize a single global reward or holistic preference signal, leaving fine-grained, attribute-level objectives such as accurate text rendering largely underexplored.

\section{Method}

\subsection{Overview of TextAlign}
\label{sec:overview}

We cast text rendering as a fine-grained alignment problem and propose \textbf{TextAlign}, a post-training preference-alignment framework dedicated to text rendering. In contrast to prior approaches that strengthen text rendering by replacing the text encoder or attaching auxiliary glyph-control modules, TextAlign introduces no additional trainable modules and leaves the network architecture of the foundation model unchanged; it sharpens typographic precision purely through preference alignment in the post-training stage. This non-invasive design offers two immediate benefits. First, the generality and diversity of the underlying image generator are preserved intact. Second, the same alignment recipe transfers across image foundation models of heterogeneous architectures without any model-specific redesign.

Formally, let $\pi_{\mathrm{ref}}$ denote the frozen reference (base) model and $\pi_\theta$ the trainable policy, with conditioning prompts $c\sim\mathcal{D}$ and samples $x\sim\pi_\theta(\cdot\mid c)$. Given a reward function $R(\cdot)$ tailored to text rendering (Sec.~\ref{sec:reward}), the TextAlign objective is

\begin{equation}
  \max_{\theta}\; \mathbb{E}_{c\sim\mathcal{D},\, x\sim\pi_\theta(\cdot\mid c)}\!\bigl[R(x, c)\bigr] \;-\; \beta\,\mathbb{D}_{\mathrm{KL}}\!\bigl[\pi_\theta\,\Vert\,\pi_{\mathrm{ref}}\bigr],
  \label{eq:align_obj}
\end{equation}

where $\beta$ controls the deviation from the reference distribution. Eq.~\eqref{eq:align_obj} is naturally compatible with mainstream preference-alignment paradigms: under GRPO~\cite{deepseekmath, dancegrpo}, the scalar reward is normalized within each sample group to obtain relative advantages; under the DPO paradigm~\cite{dpo, diffusiondpo}, the same reward is used to rank samples generated under the same condition and construct winner/loser preference pairs, which are then optimized with the Diffusion-DPO objective for diffusion-based image generators. We instantiate TextAlign with the same recipe on SD3.5, FLUX, Qwen-Image, and Z-Image to validate the architecture-agnostic nature of the framework.

\subsection{TextAlign Reward Design}
\label{sec:reward}
The crux of Eq.~\eqref{eq:align_obj} lies in constructing a reward function $R$ that faithfully reflects rendering quality. Failure modes of text rendering are inherently multi-level: from the most macroscopic case in which the image contains no readable text at all, through mid-scale errors such as dropped, inserted or substituted whole words, down to the most microscopic single-character substitutions, insertions and deletions. A reward that focuses on any single level inevitably leaves blind spots---supervising solely with OCR edit distance overlooks global structural failures, while a binary text-presence judgement cannot expose character-level defects. Furthermore, off-the-shelf OCR engines merely predict text strings and fail to evaluate visual glyph deformations, necessitating VLM-based visual inspection for structural quality.

\begin{figure*}[t]
  \centering
  \includegraphics[width=\linewidth]{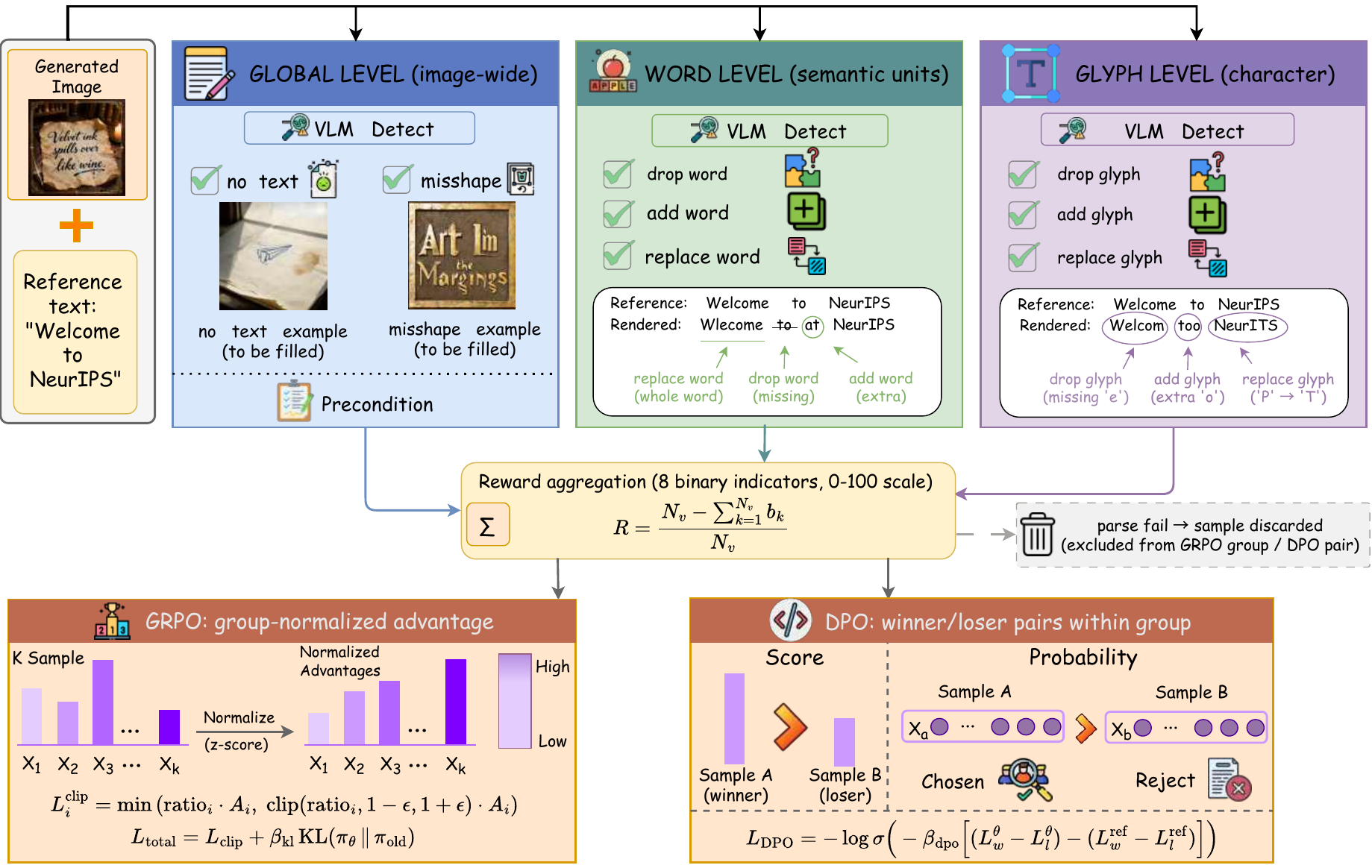}
  \caption{\textbf{Our hierarchical reward mechanism}. Given a generated image $x$ and reference text $y$, three independent VLM calls produce binary indicators at the global, word and glyph levels, which are aggregated into a scalar reward $R$ that drives either GRPO or DPO.}
  \label{fig:reward}
\end{figure*}

Motivated by this, we propose a three-level reward decomposition that covers, in a top-down fashion, the \emph{global}, \emph{word} and \emph{glyph} granularities, explicitly classifying rendering failures and dispatching each level to an independent judgement by a vision-language model (VLM). For every generated sample together with its reference text $y$, the three levels are realized as three independent VLM calls, each focusing exclusively on its own failure modes and returning a set of binary indicators. The three sets of indicators are finally compressed into a single scalar by a unified aggregation function (Sec.~\ref{sec:reward_agg}) and consumed by Eq.~\eqref{eq:align_obj}. Fig.~\ref{fig:reward} provides an overview.

\subsubsection{Global Level}

The global level addresses the most fundamental questions: whether any readable text appears in the image and whether the overall glyph forms hold up. At this level the VLM independently produces two binary indicators: $b_{\text{no text}}$ records whether the image contains no recognizable text at all, and $b_{\text{misshape}}$ records whether there exist characters of the correct identity but with severely distorted contours, including typical deformations such as broken strokes, imbalanced proportions, and warped baselines. The global level decides whether rendered content is present at all and therefore serves as a precondition for the two finer-grained levels that follow.

\subsubsection{Word Level}

The word level characterizes the model's faithfulness to semantic units. We explicitly decompose word-level failures into three atomic indicators: $b_{\text{drop word}}$ flags the presence of missing words, $b_{\text{add word}}$ the presence of extraneous words, and $b_{\text{replace word}}$ the presence of whole-word substitutions in which an intended word is rendered as a different word---typically manifested as two or more character errors occurring simultaneously. To produce these indicators, the VLM first extracts the rendered text from the image, normalizes case, removes punctuation, and performs a token-level alignment against the reference text $y$, yielding three mutually independent 0/1 signals.

\subsubsection{Glyph Level}

The glyph level targets the finest granularity of typographic precision. We symmetrically decompose character-level failures into three indicators: $b_{\text{drop glyph}}$ flags missing characters, $b_{\text{add glyph}}$ flags inserted characters, and $b_{\text{replace glyph}}$ flags single-character substitutions in which the word length is preserved. This decomposition explicitly disentangles ``length changes'' from ``content substitutions'': $b_{\text{drop glyph}}$ and $b_{\text{add glyph}}$ require the surviving characters to remain a subsequence of the reference word, whereas $b_{\text{replace glyph}}$ requires the lengths to match with exactly one differing character, thereby avoiding any conflation with the whole-word replacement detected at the word level.

\subsubsection{Reward Aggregation}
\label{sec:reward_agg}

The three levels jointly produce $N{=}8$ binary indicators $\{b_k\}_{k=1}^{N}$, each level emitted by an independent VLM call that is constrained to return a fixed-field structured response, thereby discretizing the model's qualitative judgement into parsable signals. Let $N_{\text{v}}\le N$ denote the number of indicators successfully parsed for a given sample. We define the scalar reward as:
\begin{equation}
  R \;=\; \textstyle \frac{N_{\text{v}} - \sum_{k=1}^{N_{\text{v}}} b_k}{N_{\text{v}}},
  \label{eq:reward_agg}
\end{equation}

which corresponds to the fraction of indicators reporting no defect. The aggregated scalar reward is used differently by the two optimizers: GRPO normalizes it within each group to obtain relative advantages, whereas DPO uses it only to construct winner/loser preference pairs under the same prompt. Thus, the global, word, and glyph levels require no optimizer-specific redesign; the same reward function supports both relative policy optimization and pairwise preference optimization. Whenever the structured response of any level fails to parse, we discard the entire sample and exclude it from both the group normalization in GRPO and the pair construction in DPO, in order to prevent noisy feedback from contaminating the gradients. Eq.~\eqref{eq:reward_agg} is simple in form, treats all defect categories on equal footing, and is robust to variations in the VLM output format. We adopt Qwen3.5-9B as the default reward VLM, and in Sec.~\ref{sec:exp} we further verify that the framework retains consistent performance when the reward VLM is replaced by models from the Qwen3-VL family.

\section{Experiments}
\label{sec:exp}
\subsection{Experimental Setup}
\label{sec:setup}
\textbf{Dataset construction.} Existing text rendering benchmarks are limited: prompts are typically short and templated, and target texts skew toward short, centered strings with uncontrolled semantic positions. This underrepresents real deployment difficulty, where text can be long, scenes visually complex, and target strings arbitrarily positioned. We therefore construct a more discriminative benchmark varying along three axes: text length, prompt complexity, and textual position. Samples span ten visual-text carriers differing in layout complexity, text scale, background clutter, and typographic style. See Supplementary Sec.~A for full details.

\textbf{Base models and baselines.} We conduct preference-alignment experiments across three open-source foundation models: SD3.5~\cite{sd3}, FLUX.1-dev~\cite{flux}, and Z-Image-Turbo~\cite{zimage}, applying both DPO and GRPO fine-tuning. Notably, Z-Image-Turbo is a distilled model designed for few-step generation; during alignment and evaluation, we keep its sampling steps identical to standard inference settings, showing TextAlign remains effective for fast distilled generators without extra compute overhead. We additionally include Qwen-Image~\cite{qwenimage} as a strong foundation baseline, and compare with two architecture-level text rendering methods: TextDiffuser~\cite{textdiffuser}, which guides generation via character-level segmentation masks, and AnyText~\cite{anytext}, which injects glyph and positional control via ControlNet.

\textbf{Evaluation metrics.} Our evaluation metrics are fully decoupled from the reward signals defined in Sec.~\ref{sec:reward}. For textual accuracy, we extract the rendered text from generated images using PaddleOCR-VL-1.5~\cite{paddleocrvl} and report the normalized edit score (NED; higher is better) to measure character-level similarity to the target text, together with precision, recall, F1, and accuracy computed under the word-level matching protocol of TextDiffuser~\cite{textdiffuser} to assess word-set-level agreement. For general generation quality, we report five external metrics, CLIPScore~\cite{clipscore}, Aesthetic Predictor~\cite{aesthetic}, ImageReward~\cite{imagereward}, PickScore~\cite{pickscore}, and HPSv3~\cite{hpsv3}, to verify that the alignment process does not degrade the base model's general synthesis capability.

\subsection{Main Results}
\label{sec:main-results}
Table~\ref{tab:grpo} reports OCR-based text rendering accuracy after DPO and GRPO fine-tuning. The evaluation is decoupled from the VLM reward used during training, covering the normalized edit score as well as word-level precision, recall, F1-score, and accuracy. On FLUX, both DPO and GRPO improve over the base model across all metrics, with GRPO yielding the larger gains. On the stronger Z-Image backbone, DPO also consistently improves the base model, and GRPO further achieves the best scores in all five OCR-based metrics, including exact word-set accuracy. These results suggest that the hierarchical reward is effective not only for relative policy optimization, but also for pairwise preference optimization.

Compared with architecture-level text rendering methods such as AnyText and TextDiffuser, TextAlign achieves stronger text accuracy without introducing additional modules or modifying the foundation model architecture.

\begin{table*}[t]
  \centering
  \caption{\textbf{Text rendering accuracy under DPO and GRPO fine-tuning.} We report OCR-based text accuracy metrics, including the normalized edit score, word-level precision, recall, f1-score, and exact word-set accuracy. \legendbox{rankfirst}\,Denotes the highest value in each column.}
  \label{tab:grpo}
  \resizebox{1.\linewidth}{!}{
  \renewcommand{\arraystretch}{1.05}
  \setlength{\tabcolsep}{18pt} 
  \begin{tabular}{@{}lccccc@{}}
    \toprule
    model
    & NED
    & precision & recall & f1-score & accuracy \\
    \midrule
    AnyText            & 0.2857 & 0.0014 & 0.0008 & 0.0008 & 0.0000 \\
    TextDiffuser       & 0.3766 & 0.5770 & 0.3068 & 0.3633 & 0.0852 \\
    SD3.5              & 0.5876 & 0.5242 & 0.5146 & 0.5045 & 0.1019 \\
    SD3.5(Our DPO)     & 0.5903 & 0.5356 & 0.5208 & 0.5271 & 0.1161 \\
    SD3.5(Our GRPO)    & 0.6028 & 0.5473 & 0.5385 & 0.5429 & 0.1306 \\
    Qwen-Image         & 0.8615 & 0.8563 & 0.8352 & 0.8318 & 0.4722 \\
    FLUX               & 0.5768 & 0.5746 & 0.5301 & 0.5315 & 0.1667 \\
    FLUX(Our DPO)      & 0.5925 & 0.5872 & 0.5336 & 0.5396 & 0.1778 \\
    FLUX(Our GRPO)     & 0.6075 & 0.6070 & 0.5758 & 0.5829 & 0.2037 \\
    Z-Image            & 0.8739 & 0.8860 & 0.8726 & 0.8657 & 0.5278 \\
    Z-Image(Our DPO)   & 0.8830 & 0.8994 & 0.8790 & 0.8762 & 0.5352 \\
    Z-Image(Our GRPO)  & \first{0.8893} & \first{0.9105} & \first{0.8924} & \first{0.8876} & \first{0.5648} \\
    \bottomrule
  \end{tabular}
  }
\end{table*}

\begin{table}[t]
  \centering
  \caption{\textbf{General generation quality under DPO and GRPO fine-tuning.} We report CLIPScore, ImageReward, PickScore, and HPSv3 to evaluate whether preference alignment preserves general image generation quality beyond text rendering accuracy. 
    \legendbox{rankfirst}\,1st
    \ \legendbox{ranksecond}\,2nd
    \ \legendbox{rankthird}\,3rd.}
  \label{tab:quality}
  \resizebox{\linewidth}{!}{
  \renewcommand{\arraystretch}{1.05}
  \setlength{\tabcolsep}{6pt} 
  \begin{tabular}{@{}lcccc@{}}
    \toprule
    model & CLIPScore & ImageReward & PickScore & HPSv3 \\
    \midrule
    AnyText            & 20.77 & -0.902 & 19.11 & 6.010 \\
    TextDiffuser       & 28.53 & -0.328 & 19.93 & 8.097 \\
    SD3.5              & 31.04 & 0.682 & 21.12 & 10.60 \\
    SD3.5(Our DPO)     & 31.07 & 0.680 & 21.13 & 10.62 \\
    SD3.5(Our GRPO)    & 31.09 & 0.685 & 21.12 & 10.63 \\
    Qwen-Image         & \third{31.41} & \third{0.915} & \second{21.54} & 11.65 \\
    FLUX               & 29.38 & 0.744 & 21.41 & 11.84 \\
    FLUX(Our DPO)      & 29.49 & 0.721 & 21.37 & 11.76 \\
    FLUX(Our GRPO)     & 29.65 & 0.724 & 21.41 & 11.88 \\
    Z-Image            & 31.36 & \second{0.926} & \second{21.54} & \first{12.16} \\
    Z-Image(Our DPO)   & \first{31.56} & 0.898 & \second{21.54} & \third{12.12} \\
    Z-Image(Our GRPO)  & \second{31.47} & \first{0.928} & \first{21.55} & \second{12.13} \\
    \bottomrule
  \end{tabular}
  }
\end{table}

\begin{figure}[t]
  \centering
  \includegraphics[width=\linewidth]{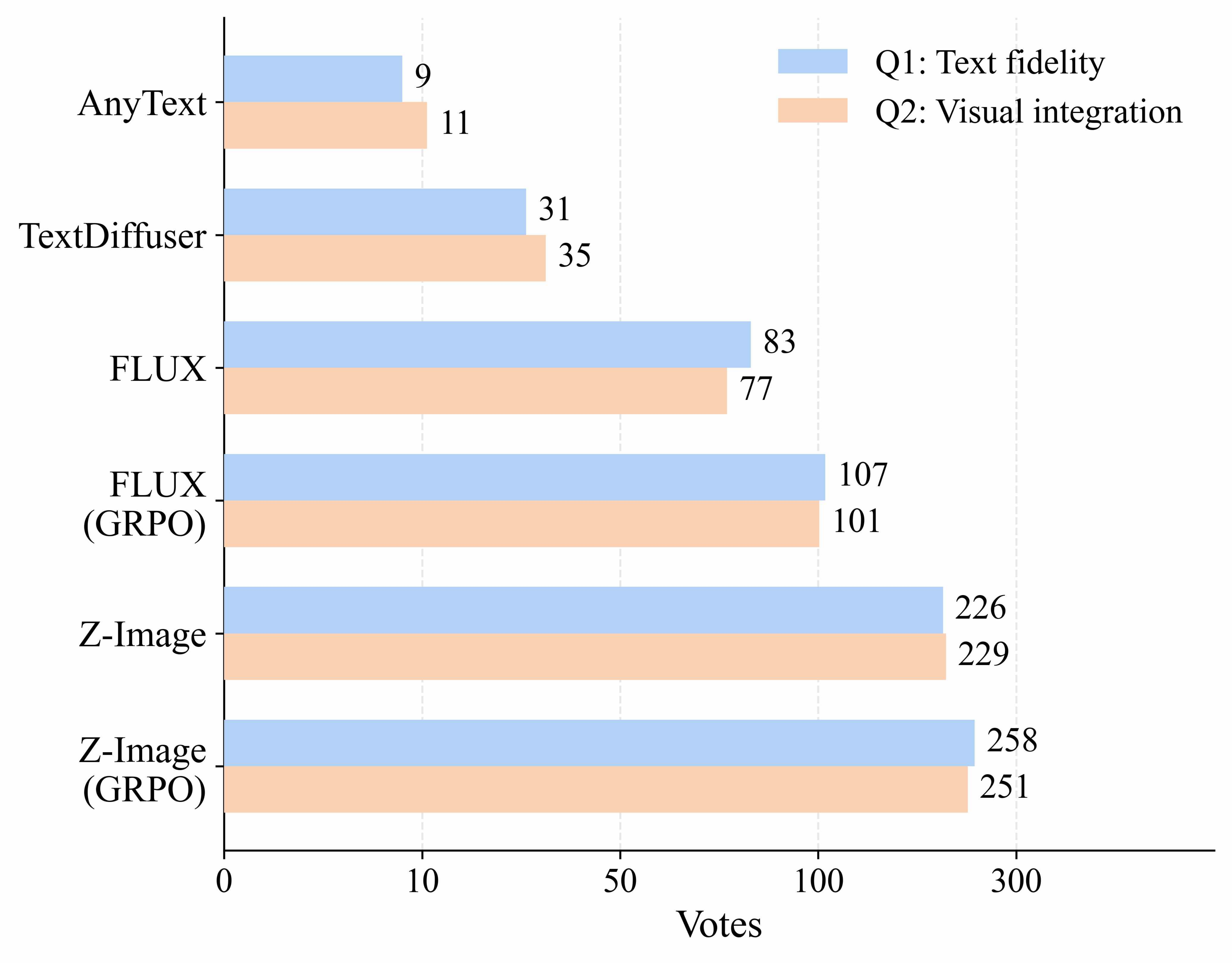}
  \caption{\textbf{User study.} Human preference votes on text fidelity and visual integration. Our GRPO-aligned models outperform prior baselines and base generators on both criteria, with Z-Image (Our GRPO) preferred most.}
  \label{fig:user-study}
\end{figure}

\begin{figure}[ht]
  \centering
  \includegraphics[width=1\linewidth]{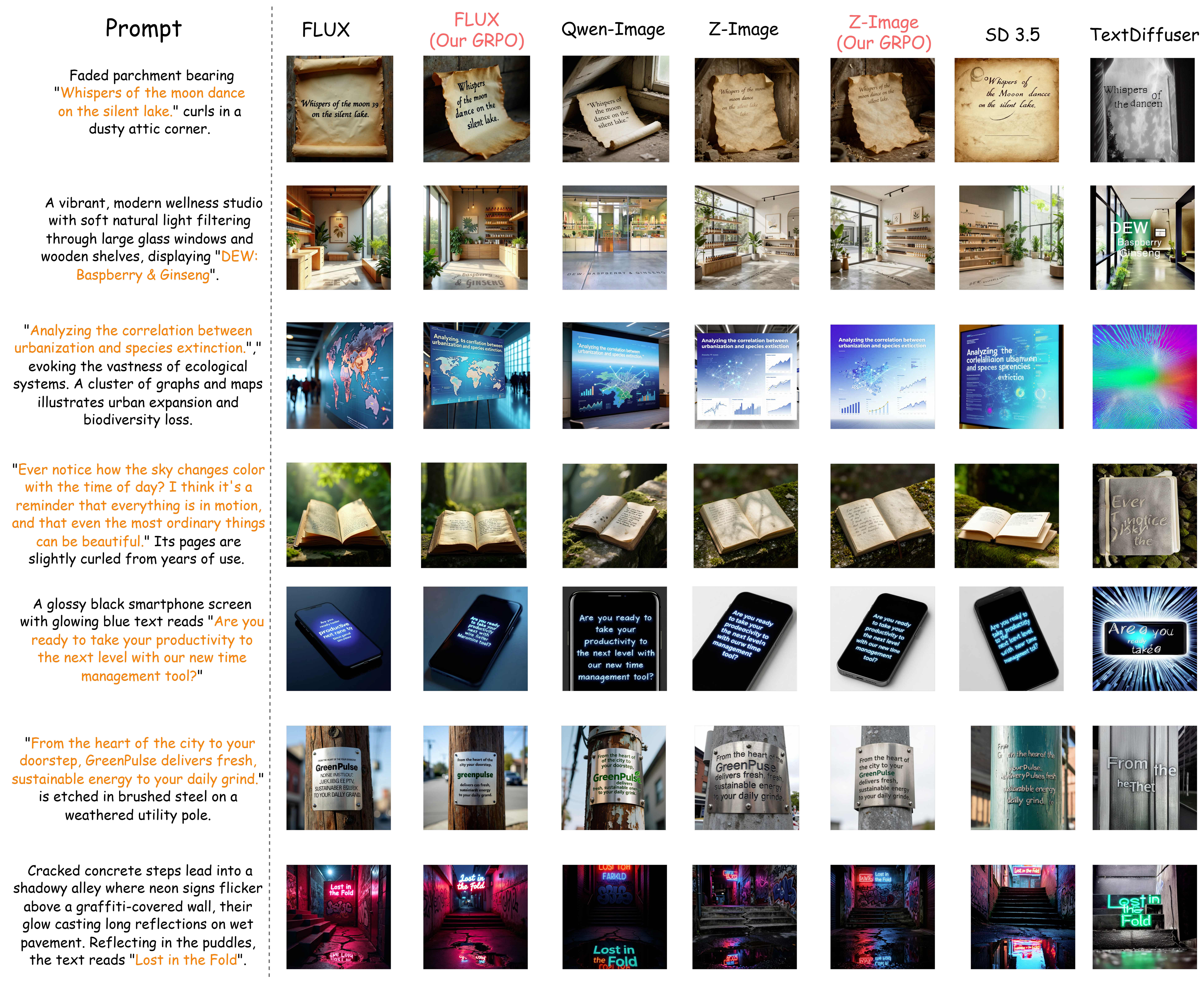}
  \caption{\textbf{Qualitative comparison of text rendering results.} Given the same prompts, GRPO-aligned FLUX and Z-Image produce more faithful and legible visual text while preserving the surrounding visual context.}
  \label{fig:results-comp}
\end{figure}

Table~\ref{tab:quality} further evaluates whether these accuracy gains preserve general generation quality. Overall, the aligned models remain competitive in CLIPScore, aesthetic quality, ImageReward, and HPSv3, and the GRPO variants maintain strong PickScore values; the main exception is FLUX (Our DPO), whose PickScore decreases despite improved text accuracy. The qualitative results in Fig.~\ref{fig:results-comp} show a consistent trend, where aligned models generate more complete, clearer, and more readable text while preserving the scene, style, and layout specified by the prompt. We further report a user study in Fig.~\ref{fig:user-study}, following the two-question protocol of TextDiffuser~\cite{textdiffuser} to evaluate text fidelity and visual integration. The aligned GRPO variants receive more votes than their corresponding base models on both criteria, with Z-Image (Our GRPO) obtaining the strongest overall preference, indicating that the OCR-based gains are also reflected in human perceptual judgments.

\subsection{Generalization Analysis}
\label{sec:generalization}
To evaluate the generalization of TextAlign across diverse visual text scenarios, we report the category-wise performance of Z-Image (Our GRPO) over the ten representative categories described above. As shown in Table~\ref{tab:category}, the model maintains strong text rendering accuracy across categories, with an average NED of 0.8848 and an average F1-score of 0.8874. The strongest results are observed on basic, handwriting, and poster, while academic and scene show relatively lower exact-match accuracy, likely due to more complex layouts, smaller text regions, and stronger background interference. Meanwhile, CLIPScore, Aesthetics, and HPSv3 remain stable across categories, suggesting that TextAlign improves text readability without noticeably weakening category-specific style or visual semantics. We further analyze robustness to text length and spatial placement in Sec.~B of the supplementary material, where the aligned model maintains stable performance across different string lengths and layout positions. Additional qualitative examples in Sec.~C of the supplementary material show that the model can naturally embed legible text into diverse carriers such as posters, logos, stickers, handwriting, and natural scenes.

\begin{table*}[t]
  \centering
  \caption{\textbf{Per-category performance of Z-Image (Our GRPO).} We report text rendering accuracy and generation quality metrics across ten visual scene categories. 
    \legendbox{rankfirst}\,1st
    \ \legendbox{ranksecond}\,2nd
    \ \legendbox{rankthird}\,3rd.}
  \label{tab:category}
  \resizebox{1.\linewidth}{!}{
  \renewcommand{\arraystretch}{1.05}
  \setlength{\tabcolsep}{8pt}
  \begin{tabular}{@{}lccccccc@{}}
    \toprule
    Category
    & NED
    & precision & recall & f1-score & accuracy
    & CLIPScore & HPSv3 \\
    \midrule
    Poster        & \third{0.9099} & 0.9281 & \third{0.9033} & \third{0.9012} & \third{0.6481} & \third{32.14} & \third{12.61} \\
    Advertisement & 0.8757 & 0.8963 & 0.8676 & 0.8651 & 0.5370 & 31.84 & 11.71 \\
    Cover         & 0.8712 & 0.9051 & 0.8924 & 0.8796 & 0.5370 & 30.00 & \second{12.63} \\
    Logo          & \first{0.9208} & 0.8960 & 0.8761 & 0.8762 & \third{0.6481} & 32.11 & 11.87 \\
    Sticker       & 0.8636 & \second{0.9359} & 0.8936 & 0.8937 & 0.5741 & \first{32.48} & 12.58 \\
    Handwriting   & \second{0.9205} & \first{0.9422} & \second{0.9485} & \first{0.9420} & \second{0.6852} & 29.89 & 11.89 \\
    Scene         & 0.8882 & 0.8886 & 0.8344 & 0.8518 & 0.4630 & 31.16 & \first{12.99} \\
    Basic         & 0.9091 & 0.9302 & \first{0.9617} & \second{0.9340} & \first{0.7222} & 31.99 & 10.99 \\
    Artistic      & 0.8776 & \third{0.9305} & 0.8753 & 0.8846 & 0.5370 & 30.75 & 12.42 \\
    Academic      & 0.8564 & 0.8518 & 0.8714 & 0.8481 & 0.2963 & \second{32.34} & 11.60 \\
    \midrule
    Average       & 0.8893 & 0.9105 & 0.8924 & 0.8876 & 0.5648 & 31.47 & 12.13 \\
    \bottomrule
  \end{tabular}
  }
\end{table*}
\subsection{Evaluation on External Benchmarks}
To test whether the gains from TextAlign transfer beyond our constructed benchmark, we evaluate the aligned models on two external benchmarks: MARIO-Eval~\cite{textdiffuser} and LongText-Bench. 

As shown in Table~\ref{tab:mario-eval}, GRPO alignment improves both evaluated backbones across all reported metrics on a 500-sample split of MARIO-Eval. For FLUX, the F1-score increases from 0.2437 to 0.2795 and accuracy rises from 0.1000 to 0.1240. For Z-Image, the aligned model similarly improves NED, precision, recall, F1-score, and exact accuracy.

We further evaluate performance on LongText-Bench~\cite{longtextbench} across English and Chinese subsets to examine long-text rendering capabilities. For LongText-Bench, to handle the increased evaluation difficulty of complex long-text scenarios, we employ a more advanced VLM, Qwen3.5-9B, as the OCR detection engine instead of Qwen2.5-VL-7B. As presented in Table~\ref{tab:longtext-bench}, GRPO alignment yields consistent performance boosts across all base models (SD3.5, FLUX.1-dev, and Z-Image-Turbo). Notably, Z-Image-Turbo (Our GRPO) achieves the highest scores in both English (0.9407) and Chinese (0.9043) text rendering. These results demonstrate that the hierarchical reward generalizes effectively to out-of-domain prompt distributions and multilingual long-text scenarios.

\begin{table}[t]
  \centering
  \caption{\textbf{Evaluation on the external MARIO-Eval benchmark.} We report OCR-based text rendering metrics on a 500-sample external split. The main comparison is between each base model and its GRPO-aligned counterpart; \legendbox{rankfirst}\,Denotes the highest value in each column.}
  \label{tab:mario-eval}
  \resizebox{\linewidth}{!}{
  \renewcommand{\arraystretch}{1.05}
  \setlength{\tabcolsep}{6pt} 
  \begin{tabular}{@{}lccccc@{}}
    \toprule
    model & NED & precision & recall & f1-score & accuracy \\
    \midrule
    AnyText                 & 0.0896 & 0.0005 & 0.0008 & 0.0004 & 0.0000 \\
    SD3.5                   & 0.3127 & 0.2621 & 0.3943 & 0.2816 & 0.1060 \\
    SD3.5(Our GRPO)         & 0.3349 & 0.2836 & 0.4217 & 0.3266 & 0.1305 \\
    FLUX                    & 0.2686 & 0.2253 & 0.3744 & 0.2437 & 0.1000 \\
    FLUX(Our GRPO)          & 0.2974 & 0.2602 & 0.4185 & 0.2795 & 0.1240 \\
    Z-Image           & 0.5130 & 0.4849 & 0.6696 & 0.5199 & 0.2460 \\
    Z-Image(Our GRPO) & \first{0.5189} & \first{0.4933} & \first{0.6852} & \first{0.5272} & \first{0.2600} \\
    \bottomrule
  \end{tabular}
  }
\end{table}

\begin{table}[!t]
  \centering
  \caption{\textbf{Evaluation on LongText-Bench.} Text rendering accuracy evaluated on English and Chinese long-text subsets. \legendbox{rankfirst}\,Denotes the highest value in each column.}
  \label{tab:longtext-bench}
  \resizebox{\linewidth}{!}{
  \renewcommand{\arraystretch}{1.05}
  \setlength{\tabcolsep}{14pt} 
  \begin{tabular}{@{}lcc@{}}
    \toprule
    Model & English & Chinese \\
    \midrule
    SD3.5                       & 0.4806 & 0.0505 \\
    SD3.5(Our GRPO)             & 0.5213 & 0.0507 \\
    FLUX                        & 0.6378 & 0.0220 \\
    FLUX (Our GRPO)             & 0.6794 & 0.0282 \\
    Z-Image-Turbo               & 0.9267 & 0.9011 \\
    Z-Image-Turbo (Our GRPO)    & \first{0.9407} & \first{0.9043} \\
    \bottomrule
  \end{tabular}
  }
\end{table}

\subsection{Ablation Study}

Table~\ref{tab:ablation-levels} ablates the hierarchical reward by removing one level at a time from the full TextAlign reward.
On FLUX.1-dev, the full reward achieves the best NED, precision, recall, and F1-score, and removing the glyph level leads to the largest drop in NED, indicating the importance of fine-grained character-level feedback.
On Z-Image, the full reward gives the strongest precision, recall, and F1-score, although some ablated variants slightly improve a single metric such as NED or strict accuracy.
Overall, the three reward levels are complementary: global feedback stabilizes readable text structure, word-level feedback preserves semantic units, and glyph-level feedback refines character accuracy.

\begin{table}[!t]
  \centering
  \caption{\textbf{Ablation on the three reward levels.} For each base model, we report the full TextAlign (all three levels) and three variants that remove the global, word, or glyph level respectively.}
  \label{tab:ablation-levels}
  \resizebox{\linewidth}{!}{
  \renewcommand{\arraystretch}{1.05}
  \setlength{\tabcolsep}{6pt} 
  \begin{tabular}{@{}cccccc@{}}
    \toprule
    model
    & NED
    & precision & recall & f1-score & accuracy\\
    \midrule
    \multicolumn{6}{c}{\textit{Base: FLUX}} \\ 
    \midrule
    FLUX(ours), full                & \first{0.6075} & \first{0.6070} & \first{0.5758} & \first{0.5829} & \first{0.1926} \\
    \quad w/o global level          & 0.5839 & 0.5799 & 0.5398 & 0.5387 & 0.1722 \\
    \quad w/o word level            & 0.5877 & 0.5810 & 0.5417 & 0.5428 & 0.1648 \\
    \quad w/o glyph level           & 0.5723 & 0.5786 & 0.5409 & 0.5356 & 0.1889 \\
    \quad FLUX (Baseline)           & 0.5768 & 0.5746 & 0.5301 & 0.5315 & 0.1667 \\
    \midrule
    \multicolumn{6}{c}{\textit{Base: Z-Image}} \\ 
    \midrule
    Z-Image(ours), full             & 0.8893 & \first{0.9105} & \first{0.8924} & \first{0.8876} & \first{0.5648} \\
    \quad w/o global level          & 0.8841 & 0.8984 & 0.8826 & 0.8772 & 0.5371 \\
    \quad w/o word level            & \first{0.8998} & 0.8803 & 0.8735 & 0.8740 & 0.5407 \\
    \quad w/o glyph level           & 0.8785 & 0.8961 & 0.8762 & 0.8742 & 0.5167 \\
    \quad Z-Image (Baseline)        & 0.8739 & 0.8860 & 0.8726 & 0.8657 & 0.5278 \\
    \bottomrule
  \end{tabular}
  }
\end{table}

\subsection{Qualitative Results and User Study}
\label{sec:qual-user}
Figure~\ref{fig:results-comp} provides qualitative comparisons under the same prompts. Compared with the base models, the GRPO-aligned FLUX.1-dev and Z-Image-Turbo models generate more complete, clearer, and more readable text while preserving the surrounding scene, style, and layout specified by the prompt. These examples show that the alignment process improves text fidelity without visibly disrupting image composition. Additional category-wise examples are provided in Sec.~C of the supplementary material. We further conduct a user study following the two-question protocol of TextDiffuser~\cite{textdiffuser}, evaluating both text fidelity and visual integration. As shown in Fig.~\ref{fig:user-study}, the aligned GRPO variants receive more human preference votes than their corresponding base models on both criteria, with Z-Image-Turbo with GRPO alignment obtaining the strongest overall preference. This confirms that the improvements measured by OCR-based metrics are also reflected in human perceptual judgments.
\section{Conclusion}
We presented \textbf{TextAlign}, a non-invasive preference-alignment framework for improving visual text rendering in large text-to-image generative models. Rather than modifying the generator architecture or adding glyph-control branches, TextAlign keeps the foundation model intact and builds a reward signal around the hierarchical structure of rendering errors. By decomposing failures into global, word, and glyph levels, the VLM-based reward distinguishes missing text, word-level mismatches, and fine-grained character defects, and supports both GRPO and DPO. Experiments on FLUX.1-dev and Z-Image-Turbo show consistent gains in OCR-based text accuracy while largely preserving general image quality. Results on an external benchmark, category-wise analysis, robustness tests over text length and placement, qualitative comparisons, and human preference evaluation further indicate that careful reward design can improve reliable text rendering without model-specific architectural redesign.

\bibliography{parts/ref}


\end{document}